\pdfoutput=1

\documentclass[table,11pt]{article}

\usepackage[preprint]{acl}

\usepackage{times}
\usepackage{latexsym}
\usepackage{arydshln}
\usepackage{newtxtext}

\usepackage[T1]{fontenc}

\usepackage[utf8]{inputenc}
\usepackage{microtype}
\usepackage{subcaption}
\usepackage{inconsolata}
\usepackage{hyperref}

\usepackage{graphicx}
\usepackage{booktabs}       
\usepackage{amsfonts}       
\usepackage{nicefrac}       
\usepackage{microtype}      
\usepackage{xcolor}         
\usepackage{xspace}
\usepackage{graphicx}
\usepackage{times}
\usepackage{latexsym}
\usepackage{multirow}
\usepackage{balance}
\usepackage{bbm}
\usepackage{mdframed}
\usepackage{paralist}
\usepackage{makecell}
\usepackage{wrapfig}
\usepackage{tabularx} 
\usepackage{amsmath}
\usepackage{amssymb}
\usepackage{mathtools}
\usepackage{amsthm}
\usepackage{lipsum}
\usepackage{multicol}
\usepackage{subcaption}
\usepackage{xcolor}
\usepackage{soul, color, xcolor}

\hypersetup{
    colorlinks,
    linkcolor={red!50!black},
    citecolor={blue!50!black},
    urlcolor={blue!80!black}
}
\usepackage{enumitem}
\usepackage{ulem}

\newcommand{\red}[1]{\textcolor{red}{#1}}
\newcommand{\green}[1]{\textcolor{green}{#1}}

\newcommand{\limegreen}[1]{\textcolor[RGB]{50,205,50}{#1}}

\definecolor{greenline}{RGB}{87,156,55}
\definecolor{blueline}{RGB}{66,117,177}
\definecolor{orangeline}{RGB}{234,137,43}

\definecolor{Blueback}{RGB}{218, 227, 243} 
\definecolor{Greenback}{RGB}{226, 240, 217}
\definecolor{Redback}{RGB}{251, 229, 214} 
\definecolor{Orangeback}{RGB}{237,102,99}
\definecolor{Grayblueback}{RGB}{127,158,188}

\newcommand{\redback}[1]{
  \begingroup
  \sethlcolor{Redback}
  \textcolor{black}{\hl{#1}}
  \endgroup
}

\newcommand{\greenback}[1]{
  \begingroup
  \sethlcolor{Greenback}
  \textcolor{black}{\hl{#1}}
  \endgroup
}

\usepackage{multicol}
\usepackage{subcaption}
\usepackage{enumitem}
\usepackage{ulem}

\newcommand{\eg}{\textit{e}.\textit{g}.}
\title{Crowd Comparative Reasoning: Unlocking Comprehensive Evaluations for LLM-as-a-Judge}

\author{Qiyuan Zhang$^{1}$\thanks{Work partially done during the internship at Huawei Noah's Ark Lab.}, Yufei Wang$^{2}$, Yuxin Jiang$^{3}$,
\textbf{Liangyou Li}$^{2}$,
\textbf{Chuhan Wu}$^{2}$, \\ \textbf{Yasheng Wang}$^{2}$, \textbf{Xin Jiang}$^{2}$, \textbf{Lifeng Shang}$^{2}$, \textbf{Ruiming Tang}$^{2}$, \textbf{Fuyuan Lyu}$^{4}$, \textbf{Chen Ma}$^{1}$\thanks{Corresponding Author.}\\
$^{1}$City University of Hong Kong,
$^{2}$Huawei Noah's Ark Lab, \\
$^{3}$The Hong Kong University of Science and Technology (Guangzhou),\\
$^{4}$McGill University \& MILA\\
\texttt{qzhang732-c@my.cityu.edu.hk}, 
\texttt{wang.yufei1@huawei.com},\\
}

\begin{document}
\maketitle
\begin{abstract}
LLM-as-a-Judge, which generates chain-of-thought (CoT) judgments, has become a widely adopted auto-evaluation method. However, its reliability is compromised by the CoT reasoning’s inability to capture comprehensive and deeper details, often leading to incomplete outcomes.
Existing methods mainly rely on majority voting or criteria expansion, which is insufficient to address the limitation in CoT. We propose Crowd-based Comparative Evaluation, which introduces additional crowd responses to compare with the candidate responses, thereby exposing deeper and more comprehensive details within the candidate responses. This process effectively guides LLM-as-a-Judge to provide a more detailed CoT judgment. Extensive experiments demonstrate that our approach enhances evaluation reliability, achieving an average accuracy gain of $6.7\%$ across five benchmarks. Moreover, our method produces higher-quality CoTs that facilitate judge distillation and exhibit superior performance in rejection sampling for supervised fine-tuning (SFT), referred to as crowd rejection sampling, thereby enabling more efficient SFT. 
Our analysis confirms that CoTs generated by ours are more comprehensive and of higher quality, and evaluation accuracy improves as inference scales. Our code is available at \url{https://github.com/Don-Joey/CCE.git}.
\end{abstract}

\section{Introduction}
\label{sec:intro}

\begin{figure}[!t]
  \centering
  \includegraphics[width=0.9\linewidth]{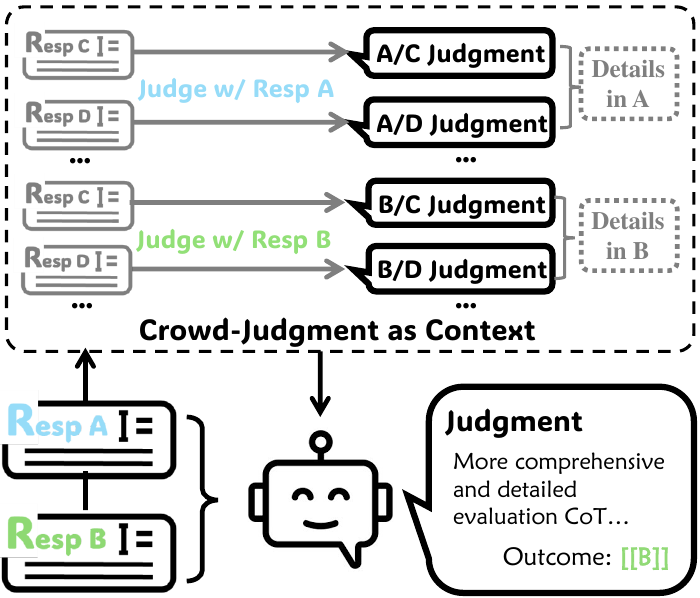}
  \caption {An overview of our method. By evaluating the candidate responses A/B alongside the crowd responses, the resulting crowd judgment can be used as context to enrich the evaluation of A/B responses, leading to a more comprehensive CoT judgment.}
  \label{fig:overview}
\end{figure}

With the prohibitive cost and limited scalability of human evaluation, LLM-as-a-Judge has emerged as a scalable framework for auto-evaluation~\citep{chang2024survey,li2024llmsas,li2025generationjudgmentopportunitieschallenges}.
Given a \textit{task instruction} and corresponding \textit{candidate responses}, LLM-as-a-Judge~\citep{zheng2023mtbench,wang2024selftaughtevaluators,wagner2024blackbox} employs CoT judgment to analyze granular quality details of the responses, ultimately deriving a final outcome.
Despite advancements in techniques such as CoT reasoning~\citep{saha2025learningplanreason,zheng2023mtbench}, specialized rubrics~\citep{liu2023geval}, and preference-aligned training datasets~\citep{li2024generative,wang2024pandalm}, human evaluation remains the gold standard due to persistent limitations~\citep{zeng2024evaluating} in LLM-as-a-Judge.
These limitations include biases~\citep{park2024offsetbias} in judgment and susceptibility to misleading context~\citep{dubois2024lengthcontrolledalpacaevalsimpleway,chen2024humans}, which undermine the reliability of automated evaluation.
One important yet overlooked reason is the quality of CoT reasoning hinges on the model’s ability to comprehensively compare nuanced details across responses. Our observation reveals high-quality judgments incorporate a thorough comparison of these details, while flawed reasoning tends to focus on limited details, leading to premature and incomplete outcomes. Therefore, enhancing the richness and comprehensiveness of CoT reasoning is essential to improve LLM-as-a-Judge.

Two commonly adopted strategies aim to address this issue: majority voting~\citep{zhang2024generative,mahan2024generativerewardmodels,deepseekai2024deepseekv3technicalreport} and criteria expansion~\citep{kim2024prometheus,liu2024hd,hu2024llmevaluator}.
The majority voting generates multiple judgments independently in parallel and aggregates these results through voting.
It essentially leverages the randomness from temperature sampling to encourage detailed reasoning.
However, this approach is passive and computationally expensive.
In contrast, criteria expansion augments prompts with additional evaluation aspects, proactively guiding the model to consider more dimensions of quality.
Yet, this strategy is response-unaware, failing to adapt the evaluation process to the unique details of each response. For instance, even if a response is rich with nuanced insights, incorporating a criterion like ``accuracy'' does little to prompt the LLM to identify the unique details of its reasoning.
Consequently, neither approach effectively guides LLM-as-a-Judge to consistently produce nuanced, comprehensive CoT evaluations.
This leads to a critical research question: \textit{how can we guide LLMs to engage in deeper, more detail-rich CoT reasoning during judgment?}

In this work, we propose a novel \textbf{crowd-based comparative evaluation} (\textbf{\textsc{CCE}}) to address this challenge by enabling LLM-as-a-Judge to uncover valuable details, as depicted in Figure \ref{fig:overview}.
Our approach is inspired by human evaluative behavior: humans merely compare candidates in isolation by also contrasting them against a broader crowd, thereby uncovering additional nuanced insights about each candidate. Building on this principle, \textsc{CCE} first gathers a set of alternative responses to the task instruction, referred to as \textit{crowd responses}, and then compares each candidate response against these crowd responses to derive multiple \textit{crowd judgments}. Throughout this process, the diversity of crowd responses serves as multiple evaluation anchors, revealing different layers of detail within the candidate responses. Based on this, \textsc{CCE} prompts the LLM-as-a-Judge to perform a more comprehensive and deeper overall CoT judgment.


CCE achieves a remarkable average improvement of $6.7\%$ across five judge benchmarks, including \textsc{RewardBench}, \textsc{HelpSteer2}, \textsc{MTBench Human}, \textsc{JudgeBench} and \textsc{EvalBias}. When applied to judge distillation, we find that the high-quality long CoT judgments generated by CCE enable a smaller judge model to achieve higher accuracy, yielding an average improvement of $4.5\%$-$5.6\%$ (in Qwen 2.5-7B), particularly enhancing bias robustness. Moreover, we extend \textsc{CCE} naturally to SFT rejection sampling, referred to as \textit{crowd rejection sampling}, where our approach serves as a quality signal to identify training-efficient samples from the response pool. Our enhanced rejection strategy consistently outperforms both random sampling and vanilla rejection sampling on \textsc{MTBench} and \textsc{AlpacaEval-v2}, demonstrating the reliability and practical utility of \textsc{CCE} in LLM alignment. Finally, our analysis confirms that \textsc{CCE} scales inference effectively and produced CoTs consistently yield more key points and capture finer-grained details within responses compared to Vanilla LLM-as-a-Judge, facilitating more comprehensive and deeper CoT reasoning. 
\section{Related Work}
\label{sec:related}
Human evaluation is typically regarded as the gold standard for evaluating LLM responses to intricate and open-ended instructions~\cite{chiang2023human, elangovan2024human}.
Nevertheless, due to its inherent limitations—being time-consuming, costly, and prone to variability~\cite{karpinska2021the}—automated evaluation methods leveraging LLMs have gained prominence as scalable and cost-efficient alternatives.
Unlike reward models that provide only scalar scores~\cite{wang2024direct,wang2024selftaughtevaluators}, LLM-as-a-Judge frameworks offer enhanced robustness and interpretability by producing detailed CoT rationales~\cite{li2024leveraginglargelanguagemodels,gao2024llmbasednlgevaluationcurrent}.

Enhancing the performance of LLM-as-a-Judge has attracted significant attention, with many techniques proposed recently.
One prominent approach involves fine-tuning pre-trained LLMs on task-specific datasets to better adapt them for judgment tasks~\cite{vu2024foundational, li2024generative, wang2024pandalm, kim2024prometheus2opensource}.
Another line of research focuses on step-by-step methodologies, such as G-EVAL~\cite{liu2023geval}, ICE-Score~\cite{zhuo2024icescore}, and EvalPlanner~\cite{saha2025learningplanreason}, which decompose complex evaluation tasks into granular components, thereby harnessing the reasoning capabilities of LLMs to streamline the evaluation process.
Additionally, recent advances explore using LLMs to generate reasoning traces by designing domain-specific prompts and meticulously crafting components of CoT reasoning.
These include constructing fine-grained scoring rubrics~\cite{zheng2023mtbench, zeng2024llmbar, trivedi2024self} and generating reference answers~\cite{zhang2025reviseval}.
Despite these efforts, the richness and comprehensiveness of CoT reasoning remain underexplored, leaving room for further advancements in improving LLM-as-a-Judge.
While simple heuristics such as majority voting~\cite{badshah2024vote, verga2024vote} can mitigate this issue by improving the reliability and accuracy of evaluations, they often fall short in terms of efficacy and efficiency.
\begin{figure*}[!t]
  \includegraphics[width=\linewidth]{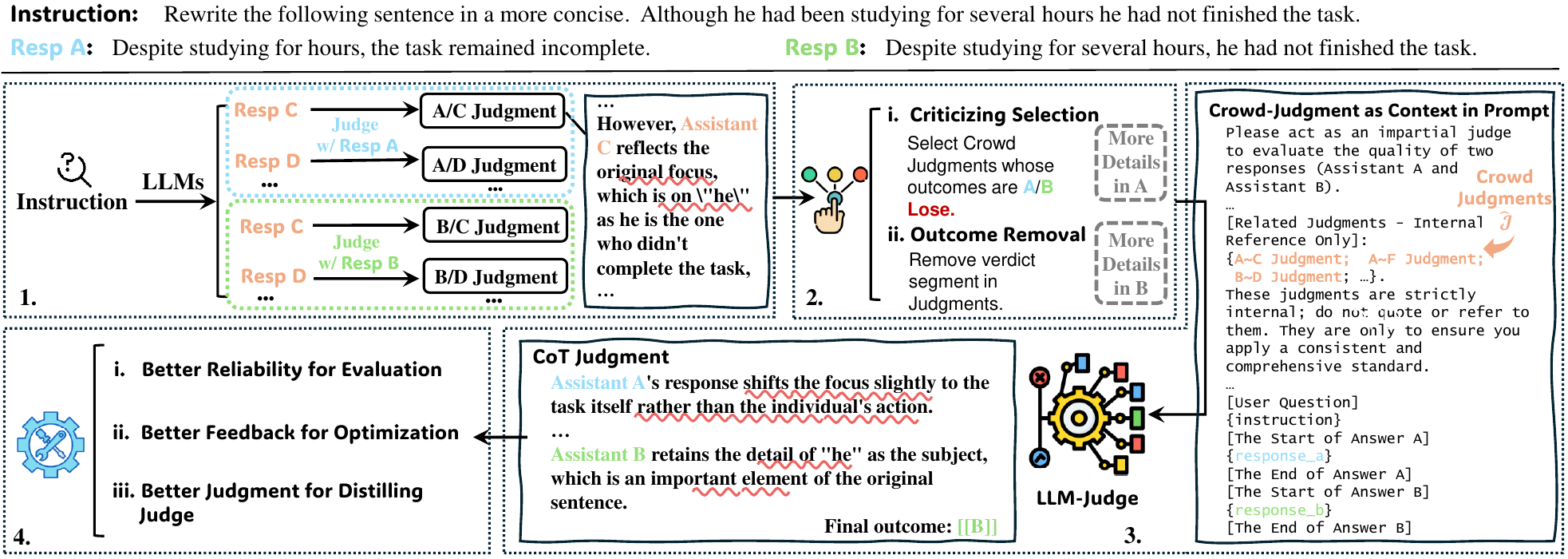}
  \caption {\textbf{Pipeline of our proposed crowd-based comparative evaluation.} For a given instance $(x, y^A, y^B)$, we first use the LLM to generate crowd responses $\left\{y^i|i\in\{C,D,E,...\}\right\}$ based on $x$. These responses are then compared with $y^A$ and $y^B$ to produce initial crowd judgments $\mathcal{J}$, which are subsequently refined into $\hat{\mathcal{J}}$ after selection and processing. Finally, $\hat{\mathcal{J}}$ are used as contextual input to evaluate the instance $(x, y^A, y^B)$.} 
  \label{fig:pipeline}
\end{figure*}

\section{Methodology}
\label{sec:method}

As illustrated in Figure~\ref{fig:pipeline}, we propose a crowd-based comparative evaluation that elicits and integrates multiple crowd judgments before producing a final outcome.
It consists of three core components: (1) Crowd Response and Judgment Generation, (2) Crowd Judgment Selection and Processing, and (3) Context-augmented Inference,  which we will discuss in the following subsections. Furthermore, we distill the CoT judgments generated by \textsc{CCE} to train a judge and expand its application to an enhanced rejection sampling technique for SFT.

\subsection{Problem Formulation}
\label{subsec:problem}
Supposing $\{y^A, y^B\}$ denote two candidate responses generated by two assistants for a given task instruction $x$, Vanilla LLM-as-a-Judge $\mathcal{F}$ is prompted to provide a CoT-based judgment $j$ of $y^A$ and $y^B$, based on a specific set of evaluation criteria $s$ 
(\eg, \textit{correctness}, \textit{coherence}).
\begin{equation}
    j = \mathcal{F}(y^A, y^B|x,s).
\end{equation}
The objective is to ensure that the $\mathcal{F}$ preference aligns closely with human evaluation. In pairwise comparisons, this alignment is quantified by measuring the accuracy relative to human labels.

\subsection{Crowd Response and Judgment Generation}
\label{subsec:crowd}

Based on the task instruction $x$, we first prompt the LLM to generate a set of $n$ synthetic crowd responses $\left\{y^i|i\in\{C,D,E,...\}\right\}$.
To enhance the diversity of these responses, we can leverage multiple LLMs ranging from smaller models (\eg, \textit{Qwen2.5-0.5B-Instruct}) to larger ones (\eg, \textit{Mistral-Nemo-Instruct-2407}), along with varying temperature settings.
Theoretically, more diverse responses can cover a wider range of scenarios. When compared with $y^A$ and $y^B$, these crowd responses emphasize different details of $\{y^A, y^B\}$, offering a more comprehensive perspective and facilitating deeper reasoning. As Figure~\ref{fig:pipeline} demonstrated, crowd judgment digs the importance of ``he'', where Response A subtly shifts the actor ``he'' onto the object ``task'' itself, thereby violating the instruction's requirement to rewrite while preserving the concise original meaning.
Then, we use it as context to reinforce the following CoT. This advantage surpasses that of criteria expansion, which cannot anticipate such details through pre-prompting.

For each synthetic $y^i$, $\mathcal{F}$ independently produces two crowd judgments, $j^A_i$ and $j^B_i$, by individually judging $y^i$ with $y^A$ and $y^B$, separately:
\begin{equation}
    j^A_i = \mathcal{F}(y^A, y^i|x,s), \quad j^B_i = \mathcal{F}(y^B, y^i|x,s).
\end{equation}
Formally, we collect a set of $2n$ crowd judgments:
\begin{equation}
    \mathcal{J} = \left\{ j^A_i, j^B_i \mid i\in\{C,D,E,...\}\right\}.
\end{equation}
While each judgment may not fully capture all details of the candidate responses, they together provide a richer pool of evidence about how $y^A$ and $y^B$ differ in nuanced ways.

\subsection{Crowd Judgment Selection and Processing}
\label{subsec:selection}
After obtaining $\mathcal{J}$, the key stage lies in selecting and processing these judgments effectively. 
Random Selection is neither stable nor optimal, so we need better strategies for using crowd judgments.

To this end, we propose a simple yet effective method called \textbf{Criticizing Selection}. Specifically, we choose judgments based on their outcomes: for $j_i^A$, we keep those where A loses, and for $j_i^B$, those where B loses. Notably, our observation reveals judgments with a critical outcome tend to provide detailed and informative reasoning for the criticized response. For instance, Judge might point out how the criticized response confuses key concepts by elaborating on specific errors in the definition and citing relevant theoretical principles. In contrast, judgments favoring the winning response tend to be brief, where the Judge might simply say, ``this answer is correct'' without further analysis. We also explore two alternative outcome-based strategies: \textbf{Praising Selection} (choosing only judgments where A/B wins) and \textbf{Balanced Selection} (maintaining an equal split between A/B wins and losses). However, as shown in our analysis (Table~\ref{tab:ablation_selection}), both strategies perform worse than Criticizing Selection. 
Additionally, to mitigate bias from the outcome distribution from crowd judgments, we introduce \textbf{Outcome Removal}, where an LLM rewrites $j_i$ to remove explicit outcome segments, ensuring a more neutral evaluation.
After the selection and processing, we obtain $\hat{{\mathcal{J}}}$. 

Notably, $j_i$ includes CoT judgments not only of the ($y^A$, $y^B$) but also of $y^i$. Our pilot study shows that removing the CoT segments about $y^i$ does not improve performance; therefore, we retain them to keep our approach simple.

\subsection{Context-augmented Inference}
\label{subsec:inference}

The final judgment is derived by evaluating responses $y^A$ and $y^B$ conditioned on the instruction $x$, the criteria $s$, and the post-processed crowd judgments $\hat{\mathcal{J}}$:
\begin{equation}
    j^\star = \mathcal{F}(y^A, y^B \mid x, s, \hat{\mathcal{J}}),
\end{equation}
where the prompt template is provided in Appendix~\ref{sec:appendix_prompt}. Notably, we distill $\{j^\star\}$ for training a smaller judge, whose performance surpasses the judge distilled from $\{j\}$, as demonstrated in Table~\ref{tab:main_distill}. It proves that higher-quality CoT judgment has better distillation efficiency.

\subsection{Extensive Application--Crowd Rejection Sampling in SFT}
\label{subsec:potential}
This subsection demonstrates the practicality of \textsc{CCE} by showcasing its extensive application in SFT.
Rejection sampling has been proven an effective augmentation technique for SFT~\citep{yuan2023scaling,zhu2023solving}. In a typical rejection sampling framework, given the task instruction and $k$ generated responses, low-quality responses are filtered out, and the remaining high-quality ones are then used for fine-tuning.
Traditionally, the Vanilla LLM-as-a-Judge selects the best response by comparing responses in pairs and choosing the one that wins most often. 
In contrast, \textsc{CCE} naturally adapts to the scenario that rejection sampling involves more than two responses, and we refer to it as \textit{crowd rejection sampling}. During pairwise comparing any two candidate responses, we effectively utilize the additional $k-2$ responses as crowd responses as introduced in Subsection~\ref{subsec:crowd}. After producing crowd judgments, it ensures a more detailed and consistent judgment.
We validate the crowd rejection sampling in our subsequent experiment (in Table~\ref{tab:main_sft}), where the integration of crowd responses consistently leads to more reliable and interpretable sampling, ultimately improving the overall performance of the fine-tuned model.
\section{Experiments}
\subsection{Experimental Setup}
We conduct a comprehensive evaluation of \textsc{CCE} across three tasks: testing preference benchmarks, judge distillation, and SFT rejection sampling. 

\begin{table*}[!t]
\centering
\small 

\resizebox{0.92\textwidth}{!}{
\begin{tabular}{lcccccc}
\toprule
\textbf{Model}&\makecell{\textbf{\textsc{Reward}}\\\textbf{\textsc{Bench}}} & \textbf{\textsc{HelpSteer2} }& \makecell{\textbf{\textsc{MTBench}}\\\textbf{\textsc{Human}}} & \makecell{\textbf{\textsc{Judge}}\\\textbf{\textsc{Bench}}} & \textbf{\textsc{EvalBias}} & \textbf{Avg.}\\

\midrule
\textbf{GPT-4o} \\
~\textit{Vanilla}&85.2&66.1&82.1&66.3&68.5&73.6\\
~\textit{LongPrompt}&86.9&67.3&81.8&63.5&70.5&74.0 \\
~\textit{EvalPlan}&88.7&65.5&81.4&62.9&74.4&74.6 \\
~\textit{16-Criteria} &87.3&69.1&82.8&66.6&73.7&75.9\\
~\textit{Maj@16} &87.9&68.9&82.4&68.6&75.5&76.7\\
~\textit{Agg@16} &88.1&68.7&82.6&67.2&77.9&76.9\\
\rowcolor{green!10}
~\textit{\textsc{CCE}-random@16} &91.2&69.5&83.1&68.9&80.1&78.6\\
\rowcolor{green!10}
~\textit{\textsc{CCE}@16} &\textbf{91.8}&\textbf{70.6}&\textbf{83.6}&\textbf{70.4}&\textbf{85.0}&\textbf{80.3}\\
\midrule
\textbf{Qwen 2.5 7B-Instruct} \\
~\textit{Vanilla}&78.2&60.7&76.1&58.3&57.4&66.1\\
\rowcolor{green!10}
~\textit{\textsc{CCE}@16}&\textbf{80.4}&\textbf{64.2}&\textbf{76.7}&\textbf{64.0}&\textbf{79.4}&\textbf{72.9}\\
\midrule
\textbf{Qwen 2.5 32B-Instruct} \\
~\textit{Vanilla}&87.4&\textbf{72.3}&79.0&68.9&71.1&75.7\\
\rowcolor{green!10}
~\textit{\textsc{CCE}@16}&\textbf{90.8}&72.1&\textbf{82.1}&\textbf{70.6}&\textbf{80.5}&\textbf{79.2}\\
\midrule
\textbf{Qwen 2.5 72B-Instruct} \\
~\textit{Vanilla}&85.2&\textbf{69.5}&79.5&68.3&68.5&74.0\\
\rowcolor{green!10}
~\textit{\textsc{CCE}@16}&\textbf{93.7}&68.5&\textbf{88.9}&\textbf{75.7}&\textbf{85.9}&\textbf{82.7}\\
\midrule
\textbf{Llama 3.3 70B-Instruct} \\
~\textit{Vanilla}&86.4&70.4&81.1&67.1&70.6&75.1\\
\rowcolor{green!10}
~\textit{\textsc{CCE}@16}&\textbf{91.7}&\textbf{71.3}&\textbf{83.5}&\textbf{69.7}&\textbf{79.2}&\textbf{79.1}\\
\bottomrule
\end{tabular}
}
\caption{Accuracy of LLM-as-a-Judge on pair-wise comparison benchmarks. \textsc{CCE} can consistently enhance the LLM-as-a-Judge's performance across 5 benchmarks, especially considerably outperforming other scaling inference strategies, like maj@16. The highest values are \textbf{bolded}. Here, \textit{\textsc{CCE}-random} refers to replacing the ``Criticizing Selection$+$Outcome-Removal Processing'' with ``Random Selection''.
}
\label{tab:main_preference}
\end{table*}

\paragraph{Preference Benchmarks and Baselines.} We adopt 5 preference benchmarks to test LLM-as-a-Judge, including \textsc{RewardBench}~\citep{lambert2024rewardbench}, \textsc{HelpSteer2}~\citep{wang2024helpsteer}, \textsc{MTBench-Human}~\citep{zheng2023mtbench}, \textsc{JudgeBench}~\citep{tan2025judgebench}, and \textsc{EvalBias}~\citep{park2024offsetbias}. These benchmarks provide general instructions across a wide range of tasks with diverse responses and use accuracy to measure their evaluation performance. They each focus on different aspects. For example, \textsc{RewardBench} covers a wider range of scenarios, while \textsc{EvalBias} focuses on various bias scenarios. We verify the generality of \textsc{CCE} on 5 LLMs and compare it against multiple baselines. In particular, we consider \textbf{Vanilla}, which uses the general LLM-as-a-Judge prompt implemented by \textsc{RewardBench}; \textbf{Maj@16}, where we independently judge a case 16 times and take a majority vote of the outcomes; \textbf{Agg@16}, where instead of majority voting, the 16 individual judgments are fed back into the LLM to aggregate a final decision; \textbf{16-Criteria}, which incorporates 16 criteria with corresponding descriptions in the prompt as designed in~\citet{hu2024arellm} and~\citet{wang2024helpsteer}; \textbf{LongPrompt}, where the LLM is explicitly directed to produce a longer CoT; and \textbf{EvalPlan}, in which an unconstrained evaluation plan is first generated based on the target case and then executed to derive the final judgment~\citep{saha2025learningplanreason}. Additional details on the preference benchmarks and baselines can be found in Appendix~\ref{sec:testing}.

\paragraph{Distilling CoT for Training Judge.} We start with a large preference dataset and evaluate it using the Vanilla LLM-as-a-Judge and \textsc{CCE} under \textit{GPT-4o-as-a-Judge}, producing two CoTs. We then pair each CoT with the original preference data to form two separate training sets, which we use to fine-tune a smaller LLM as a judge. The resulting judges’ performance clearly reflects the quality and effectiveness of each CoT. We use \textbf{TULU3-preference} data as the distillation query while the preference benchmarks for evaluating the judge remain the same as previously introduced. Details of the training implementation are provided in Appendix~\ref{sec:distilling4training}.

\paragraph{SFT Rejection Sampling.} Firstly, we generate a pool of 4 responses based on a given task instruction to serve as the rejection sampling base. We compare Crowd Rejection Sampling against Random Selection and a Vanilla Rejection Sampling method to select the best response for fine-tuning.

We select two datasets of different scales, \textbf{LIMA}~\citep{zhou2023lima} ($1$K) and \textbf{TULU3-SFT}~\citep{lambert2025tulu3} (sample $10$K), as instruction query. \textit{GPT-4o} served as the judge LLM, while \textit{Llama-3.1-8B} and \textit{Qwen-2.5-7B} are used as base models for SFT. We then evaluate the generative ability of finetuned models using \textsc{MTBench} and \textsc{AlpacaEval-2}~\citep{dubois2024lengthcontrolled}. Details of the implementation are provided in Appendix~\ref{sec:sft_data_selection}.

\begin{table*}[!t]
\centering
\small 
\resizebox{0.96\textwidth}{!}{
\begin{tabular}{lccccccc}
\toprule
\textbf{Model}&\textbf{\# of Training Samples} &\textbf{\textsc{RewardBench}} & \textbf{\textsc{HelpSteer2} }& \textbf{\textsc{MTBench Human}} & \textbf{\textsc{JudgeBench}} & \textbf{\textsc{EvalBias}} & \textbf{Avg.}\\
\midrule
\textbf{JudgeLM-7B}~\citep{zhu2023judgelmfinetunedlargelanguage}&100,000&\underline{46.4}&\underline{60.1}&64.1&32.6&\textbf{42.4}&\underline{49.1}\\
\textbf{PandaLM-7B}~\citep{wang2024pandalm}&300,000&45.7&57.6&\underline{75.0}&36.0&27.0&48.3\\
\textbf{Auto-J-13B}~\citep{li2024generative}&4,396&\textbf{47.5}&\textbf{65.1}&\textbf{75.2}&\textbf{50.9}&16.5&\textbf{51.0}\\
\textbf{Prometheus-7B}~\citep{kim2024prometheus}&100,000&34.6&30.8&52.8&9.3&11.7&27.8\\
\textbf{Prometheus-2-7B}~\citep{kim2024prometheus2opensource} &300,000&43.7&37.6&55.0&\underline{39.4}&\underline{39.8}&43.1\\
\midrule
\textbf{Llama-3.1-8B-Tuned} &&&&&&&\\
~\textit{Synthetic Judgment from Vanilla}&10,000&66.8&56.0&71.6&\underline{60.1}&34.2&57.7\\
~\textit{Synthetic Judgment from Vanilla}&30,000&\textbf{72.5}&\underline{58.6}&\underline{73.9}&50.4&\underline{46.2}&60.3\\
~\textit{Synthetic Judgment from \textsc{CCE}}&10,000&69.7&\underline{58.6}&72.7&\textbf{66.4}&38.7&\textbf{61.2}\\
~\textit{Synthetic Judgment from \textsc{CCE}}&30,000&\underline{70.0}&\textbf{60.1}&\textbf{74.3}&50.3&\textbf{50.7}&\underline{61.1}\\
\midrule
\textbf{Qwen 2.5-7B-Tuned} &&&&&&&\\
~\textit{Synthetic Judgment from Vanilla}&10,000&68.1&55.6&70.7&\underline{50.2}&38.4&56.6\\
~\textit{Synthetic Judgment from Vanilla}&30,000&71.4&56.2&75.1&48.2&54.7&61.1\\
~\textit{Synthetic Judgment from \textsc{CCE}}&10,000&68.8&56.7&71.3&49.8&40.2&57.4\\
~\textit{Synthetic Judgment from \textsc{CCE}}&30,000&\underline{73.3}&\underline{59.5}&\underline{74.9}&50.1&\underline{57.1}&\underline{63.0}\\
~\textit{Mix Synthetic Judgment from \textsc{CCE}\&Vanilla}&60,000&\textbf{74.1}&\textbf{60.7}&\textbf{76.6}&\textbf{61.6}&\textbf{60.6}&\textbf{66.7}\\
\bottomrule
\end{tabular}
}
\caption{Accuracy of Trained small LLM-as-a-Judge on pair-wise comparison benchmarks. Under the same preference pairs data, the model trained with judgments synthesized using \textsc{CCE} achieves more reliable evaluation results. The highest values are \textbf{bolded}, and the second highest is \underline{underlined}.}
\label{tab:main_distill}
\end{table*}

\subsection{Experiment Result}
In this section, we present our main results. The preference benchmark results are shown in Table~\ref{tab:main_preference}, the efficacy of distilling CoT for training smaller judges is summarized in Table~\ref{tab:main_distill}, and the training efficiency of SFT rejection sampling is reported in Table~\ref{tab:main_sft}. These three objectives are concluded across various judge LLMs and downstream tasks. Our findings for each task are as follows.

\paragraph{Performance on Preference Benchmarks.} Table~\ref{tab:main_preference} highlights \textbf{\textsc{CCE} consistently achieves state-of-the-art performance across all preference benchmarks}. First, it outperforms the Vanilla LLM-as-a-Judge, which already demonstrates reasonable reliability on multiple LLMs and benchmarks. Notably, with \textit{Qwen 2.5-72B-Instruct} as the judge, our method achieves an $8.5$ increase on \textsc{RewardBench} and an overall average gain of $8.7$. 

Second, \textbf{\textsc{CCE} proves considerably more effective than common scaling strategies such as \textit{Maj@16} and 16-Criteria}. Even with random selection, \textit{Maj@16} underperforms \textsc{CCE} by an average of 1.9. Although \textit{EvalPlan} offers a more response-aware reasoning process than \textit{16-Criteria}, its effectiveness remains lower $2.0$-$3.7$ than \textsc{CCE}. Simply generating longer CoT also falls short, indicating that scaling inference-time computation calls for a more nuanced approach.

\begin{table}[!thbp]
  \centering
  \resizebox{0.45\textwidth}{!}{
  \begin{tabular}{lcc}
    \hline
    \textbf{Rejection Sampling Method} & \textbf{\textsc{MTBench}} & \textbf{\textsc{AlpacaEval-2}} \\
    \midrule
    \multicolumn{3}{c}{Llama 3.1 8B Base} \\
    \midrule
    \textbf{Instructions from LIMA \# 1K}&&\\
    ~\textit{Random Sampling} &\underline{4.33}&2.89/3.29 \\
    ~\textit{Vanilla Rejection Sampling} &4.28&\underline{2.91/3.29} \\
    ~\textit{Crowd Rejection Sampling} &\textbf{4.53}&\textbf{3.02/3.31} \\
    \textbf{Instructions from Tulu 3 \# 10K}&&\\
    ~\textit{Random Sampling} &7.51&12.81/12.45 \\
    ~\textit{Vanilla Rejection Sampling}&\underline{7.56}&\underline{19.92/17.17} \\
    ~\textit{Crowd Rejection Sampling} &\textbf{7.63}&\textbf{22.23/19.74} \\
    \midrule
    \multicolumn{3}{c}{Qwen 2.5 7B Base} \\
    \midrule
    \textbf{Instructions from LIMA \# 1K}&&\\
    ~\textit{Random Sampling} &\underline{8.06}&\underline{14.52/9.40}\\
    ~\textit{Vanilla Rejection Sampling} &7.91&14.40/9.44  \\
    ~\textit{Crowd Rejection Sampling} &\textbf{8.63}&\textbf{14.86/9.59}\\
    \textbf{Instructions from Tulu 3 \# 10K}&&\\
    ~\textit{Random Sampling} &8.36&21.39/13.68 \\
    ~\textit{Vanilla Rejection Sampling} &\textbf{8.46}&\underline{22.71/16.44} \\
    ~\textit{Crowd Rejection Sampling} &\underline{8.41}&\textbf{23.78/17.56}  \\
    
    \bottomrule
  \end{tabular}
  }
  \caption{SFT Rejection Sampling Performance on the Instruction-Following Benchmark.
  The model fine-tuned with responses sampled using \textsc{CCE} demonstrates improved generative performance.}
  \label{tab:main_sft}
\end{table}

\begin{table*}[!tp]
\centering
\small 

\resizebox{0.96\textwidth}{!}{
\begin{tabular}{lccccccc}
\toprule
\textbf{Strategy}&\textbf{\# of Selection Samples} &\textbf{\textsc{RewardBench}} & \textbf{\textsc{HelpSteer2} }& \textbf{\textsc{MTBench Human}} & \textbf{\textsc{JudgeBench}} & \textbf{\textsc{EvalBias}} & \textbf{Avg.}\\

\midrule
~\textit{Random-Selection} &8&91.0&\underline{69.9}&82.6&68.7&78.4&78.1\\
~\textit{Praising-Selection} &8&86.6&64.2&81.5&67.1&77.7&75.4\\
~\textit{Criticizing-Selection} &8&\underline{91.2}&69.2&\underline{83.0}&68.9&79.1&78.3\\
~\textit{Balanced-Selection} &8&90.7&68.6&82.8&67.4&78.7&77.6\\
~\textit{Outcome-Removal Random-Selection} &8&\textbf{91.5}&\underline{69.9}&\underline{83.0}&\underline{69.4}&\underline{79.5}&\underline{78.7}\\
~\textit{Outcome-Removal Criticizing-Selection (Sota)} &8&\textbf{91.5}&\textbf{70.1}&\textbf{83.2}&\textbf{69.5}&\textbf{79.9}&\textbf{78.8}\\
\midrule
~\textit{Random-Selection} &16&91.2&69.5&83.1&68.9&80.1&78.6\\
~\textit{Praising-Selection} &16&87.0&68.4&82.0&67.1&77.9&76.5\\
~\textit{Criticizing-Selection} &16&90.8&\underline{69.7}&83.0&69.6&\underline{82.9}&\underline{79.2}\\
~\textit{Balanced-Selection} &16&90.6&69.3&82.9&68.0&79.6&78.1\\
~\textit{Outcome-Removal Random-Selection} &16&\underline{91.7}&\underline{69.7}&\underline{83.2}&\underline{70.0}&81.5&\underline{79.2}\\
~\textit{Outcome-Removal Criticizing-Selection(Sota)} &16&\textbf{91.8}&\textbf{70.6}&\textbf{83.6}&\textbf{70.4}&\textbf{85.0}&\textbf{80.3}\\

\bottomrule
\end{tabular}
}
\caption{Accuracy of \textsc{CCE} using different selection strategies on LLM-as-a-Judge benchmarks. Our proposed \textit{Outcome-Removal Criticizing-Selection} consistently surpasses performances using other selection strategies during the test-time inference phase.}
\label{tab:ablation_selection}
\end{table*}

\begin{figure*}[h]
\centering
  \includegraphics[width=0.96\linewidth]{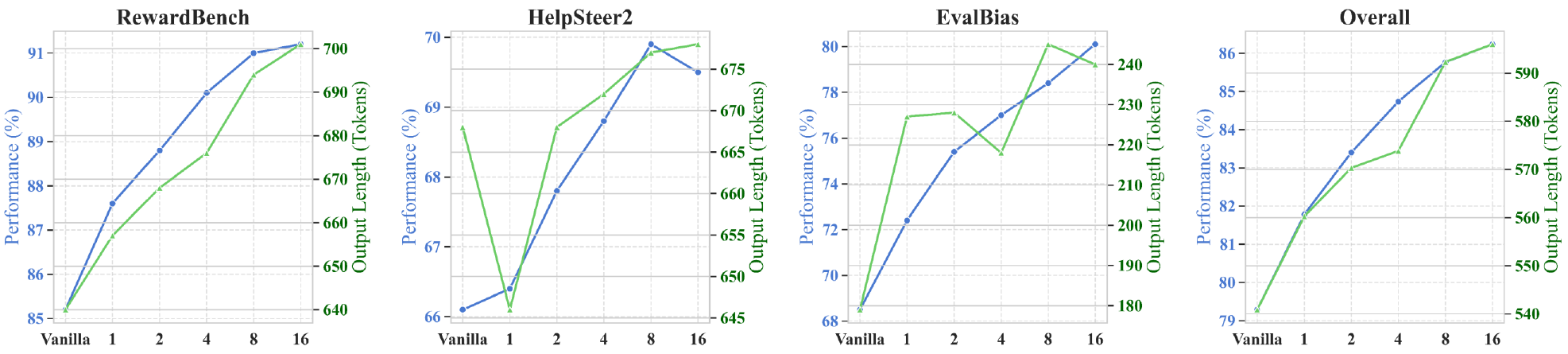}
  \caption {Evaluation performance under scaling crowd judgments in the context. As the number of crowd judgments grows, both accuracy and CoT length generally increase.}
  \label{fig:scaling}
\end{figure*}

Finally, \textsc{CCE} not only excels on \textsc{RewardBench}, the most general benchmark, but also \textbf{outperforms alternatives on more challenging tasks} like \textsc{JudgeBench} and \textsc{EvalBias}. Strategic crowd judgment selection further enhances performance compared to random selection. We adopt a ``Criticizing Selection + Outcome Removal'' strategy for our SOTA selection \& processing strategy, which we discuss in detail in the following analysis.

\paragraph{Distilling CoT for Training Smaller Judges.} Distilling preference evaluation capabilities from powerful LLMs to train smaller LLMs is a promising direction. Table~\ref{tab:main_distill} demonstrates that higher-quality CoT leads to more effective distillation, resulting in improved performance for smaller judge models. Fine-tuning small models (\eg, \textit{Llama 3.1-8B} and \textit{Qwen 2.5-7B}) on the CoTs generated by \textsc{CCE} yields higher accuracy on all five benchmarks than using \textit{Vanilla} CoTs. For instance, \textit{Qwen 2.5-7B} trained on \textsc{CCE}'s synthetic CoT judgments achieves up to 73.3\% on \textsc{RewardBench}, surpassing Vanilla baseline by a notable margin of 1.9. Moreover, combining both \textit{Vanilla} and \textsc{CCE} synthetic judgments further boosts performance, reaching 74.1\% on \textsc{RewardBench} and 60.6\% on \textsc{EvalBias}. This result suggests integrating diverse CoT can further enhance accuracy and generalization.

LLM-as-a-Judge can develop biases in various scenarios, such as favoring more verbose answers. This issue is particularly pronounced in smaller judge models. As shown in Table~\ref{tab:main_distill}, even after fine-tuning on over 100K samples, many baseline models struggle to exceed 50\% accuracy. This highlights the persistent challenge of evaluation bias. \textbf{Higher-quality and more comprehensive CoT distillation enhances the debiasing ability of smaller judge models}. These findings suggest that many biases stem from the model focusing on limited aspects of the responses rather than assessing them holistically.

\paragraph{Efficacy in SFT Rejection Sampling.} As we can see in Table~\ref{tab:main_sft}, Crowd Rejection Sampling proves effectiveness for both $1$K and $10$K data sizes, consistently \textbf{yielding better finetuning performances for two base LLMs}. \textsc{CCE} selects higher-quality responses compared to both Random Sampling and Vanilla Rejection Sampling, leading to consistent improvements in downstream instruction-following benchmarks on \textsc{MTBench} and \textsc{AlpacaEval-2}. For instance, with \textit{Llama 3.1-8B} and the TULU3-SFT instructions, the fine-tuned model sees performance gains of up to $22.23$/$19.74$ on \textsc{AlpacaEval-2}, compared to $19.92$/$17.17$ under the Vanilla Rejection Sampling. This underscores the reliability of \textsc{CCE} in identifying higher-quality training examples.

Overall, the experiments confirm the flexibility and effectiveness of \textsc{CCE} in three key general scenarios. By \textbf{leveraging crowd-based context, scaling inference-time computation, and strategically guiding the CoT process}, \textsc{CCE} delivers consistent improvements over strong baselines.

\subsection{Analysis Experiments}
In this section, we conduct an in-depth analysis of the two core components of our method: crowd judgment selection \& processing strategies, as well as inference scaling. We then directly examine whether the generated CoT is more comprehensive and provides a more detailed analysis of the responses under evaluation.

\paragraph{Selection \& Processing Strategy.}
We compare Random Selection, Criticizing Selection, Praising Selection, and Balanced Selection.
As shown in Table~\ref{tab:ablation_selection}, Criticizing Selection yields the best results, followed by Balanced Selection, while Praising Selection performs even worse than Random Selection. This suggests that \textbf{lose-based judgments provide deeper insights into A/B comparisons, making criticism more informative}. Additionally, the \textbf{Outcome-Removal post-processing strategy substantially improves evaluation reliability}, likely because final verdicts lack valuable details while introducing biases into LLM decision-making.

\paragraph{Inference Scaling.} 
Figure~\ref{fig:scaling} illustrates our analysis of how scaling crowd judgments influence evaluation outcomes. Measuring accuracy and the average token length of the CoT, three preference benchmarks are tested across different judgment counts and then averaged for an overall assessment. The implementation details are in Appendix~\ref{sec:infer_scal_appendix}.

As shown in Figure~\ref{fig:scaling}, \textbf{both performance and output length generally increase as crowd judgments rise from 0 to 16}. \textsc{RewardBench} displays a clear upward trend, while \textsc{HelpSteer2} dips briefly at 2 judgments before recovering. Averaging across benchmarks (rightmost panel) confirms that more crowd judgments lead to higher accuracy and longer CoT, consistent with the inference scaling observed in studies~\citep{brown2024largelanguagemonkeysscaling,snell2025scaling}.
Furthermore, we reexamine the Table~\ref{tab:main_preference} and find that \textbf{scaling test-time inference is a promising strategy for LLM-as-a-Judge}, as demonstrated by \textit{GPT-4o-as-a-Judge}. This is especially evident in bias scenarios, where the Vanilla struggles, while scaling-inference-based baselines, including \textsc{CCE}, show substantial gains.

\begin{figure}[t]
\centering
  \includegraphics[width=0.96\linewidth]{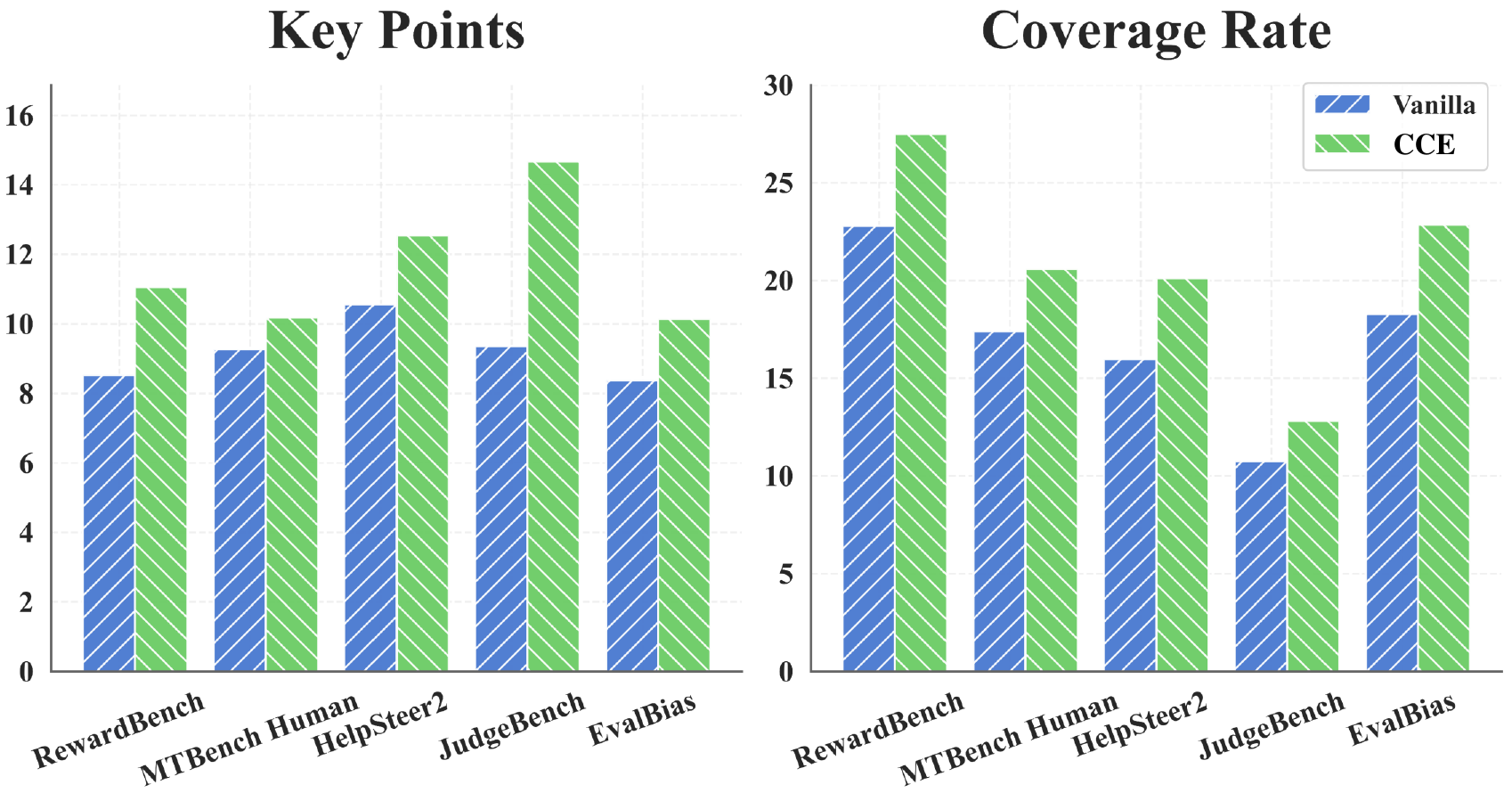}
  \caption {CoT Comparison. \textsc{CCE}’s CoT consistently yields a higher average number of key points and a higher coverage rate across all benchmarks.}
  \vspace{-.5em}
  \label{fig:cot_comp}
\end{figure}

\paragraph{CoT Comparison.} To more directly assess whether the CoTs generated by \textsc{CCE} are more comprehensive than those of the Vanilla approach, we perform two analyses: Key Points Counting and Coverage Rate. First, we use \textit{GPT-4o} to parse and summarize each CoT, counting the key points to measure how thoroughly the CoT is. Second, we leverage the cross-attention mechanism from \textit{Bart-base} to quantify the coverage rate--how thoroughly CoT covers details in the candidate responses. We introduce the details in the Appendix~\ref{subsec:coverage_appendix}.

As shown in Figure~\ref{fig:cot_comp}, \textsc{CCE} outperforms Vanilla across all benchmarks in key point counting and coverage rate. More key points indicate that our CoT examines the text from multiple angles, while a higher coverage rate reflects a more detailed analysis. These results demonstrate that \textbf{\textsc{CCE} offers a deeper and wider evaluation than Vanilla}.

\begin{table}[!ht]
  \centering
  \scriptsize
  \resizebox{0.46\textwidth}{!}{
    \begin{tabular}{p{26.5em}}
    \toprule
    \textbf{Instruction}: Classify this tweet from the categories of positive, negative, or neutral. Tweet: I'm so thankful for all the wonderful people in my life \\
    \midrule
    \textbf{\red{Response A}} \\
        \ \ \ \ The sentiment of this social media comment is positive. The use of the word "grateful" implies appreciation and thankfulness, while "amazing experiences" suggest that these experiences have been very fulfilling or enjoyable. \\
    \textbf{\limegreen{Response B}} \\
            \ \ \ \ Positive\\
    \midrule
    \textbf{Vanilla LLM-as-a-Judge}:\\

	\ \ \ \ Assistant A's . . . . . . \redback{It highlights specific words and phrases in the tweet, such as ``thankful'' and ``wonderful people,'' to justify the positive sentiment classification. This explanation adds depth and clarity to the response, making it more helpful and informative.}

    Assistant B, . . . . . .  the response less informative and helpful . . . 

    \ \ \ \ . . . . . .

    \ \ \ \ - Final Verdict: \red{[[A]]}
    \\
    \midrule
    \textbf{\textsc{CCE}}:\\

    \ \ \ \ \textbf{\textit{AC Judgment}}: \uwave{However, Assistant A makes a mistake by referencing words not present in the tweet, such as "grateful" and "amazing experiences,"} . . . . . . Assistant C also classifies the tweet as positive and provides a detailed explanation . . .

    \ \ \ \

    \ \ \ \ . . . . . . Assistant A . . . . .\greenback{, but it inaccurately references words not present in the tweet, such as "grateful" and "amazing experiences." This detracts from the accuracy of the response and could potentially confuse the user.} . . . . . .
    
    \ \ \ \ Assistant B \greenback{is concise and correctly classifies the tweet as positive. However, it lacks any explanation or reasoning, which limits its helpfulness and depth.} . . . . . .

    \ \ \ \ In comparing the two, \greenback{Given the importance of accuracy and explanation in sentiment analysis,} . . . . . .

    \ \ \ \ - Final Verdict: \green{[[B]]}
    \\
    \bottomrule
    \end{tabular}%
    }
  \caption{A pairwise comparison case evaluated by different methods. \limegreen{Preference} refers to right result and \red{Preference} refers to wrong result. We emphasize the noisy evaluation elements in \redback{orange}, while highlighting the useful elements of the evaluation in \greenback{limongreen}.}
  \label{tab:case-evaluation-simple}%
\vspace{-.5em}
\end{table}%

\paragraph{Case Study.} Table~\ref{tab:case-evaluation-simple} presents a representative case. The vanilla is misled by fake information in Response A, causing it to overlook the Instruction and mistakenly rate Response A as more helpful. In contrast, the crowd judgment correctly identifies the error in Response A and informs subsequent evaluations. Additionally, our method produces a more detailed CoT thereby enriching the overall evaluation process, as evidenced by statements like ``Assistant A does provide a brief explanation''.

\section{Conclusion}
\label{sec:conclusion}
In this work, we tackle the shortcomings of LLM-as-a-Judge, which stem from CoT reasoning lacking comprehensiveness and detail, by drawing inspiration from human evaluative behavior. We introduce a novel crowd-based comparative evaluation framework that enriches the CoT process to unlock more comprehensive and reliable evaluations. By scaling inference more effectively, our method serves as an efficient alternative to traditional majority voting and criteria expansion. Importantly, we demonstrate that high-quality CoT judgments boost evaluation reliability and distilling efficiency across multiple benchmarks, while broadening the scope of crowd-based evaluation applications.

\section*{Limitations}
\paragraph{Progressive Self-Iteration Paradigm.} A limitation of our work is that we do not explore self-iteration in this study, despite its potential for enhancing the evaluation process. Our method inherently allows for iterative refinement, which could be further extended into a progressive paradigm. We leave this direction for future work, aiming to investigate how iterative self-improvement can further enhance evaluation quality and robustness.

\paragraph{Selection based on LLMs.} We identify that the quality of crowd judgments influences the CoT and explore a simple yet efficient selection strategy. We generate crowd responses using many LLMs, but we do not explore which LLM's crowd response has a greater influence on crowd judgment.

\bibliography{custom}

\appendix

\section{Prompt Template}
\label{sec:appendix_prompt}

\begin{figure*}[!t]
  \includegraphics[width=\linewidth]{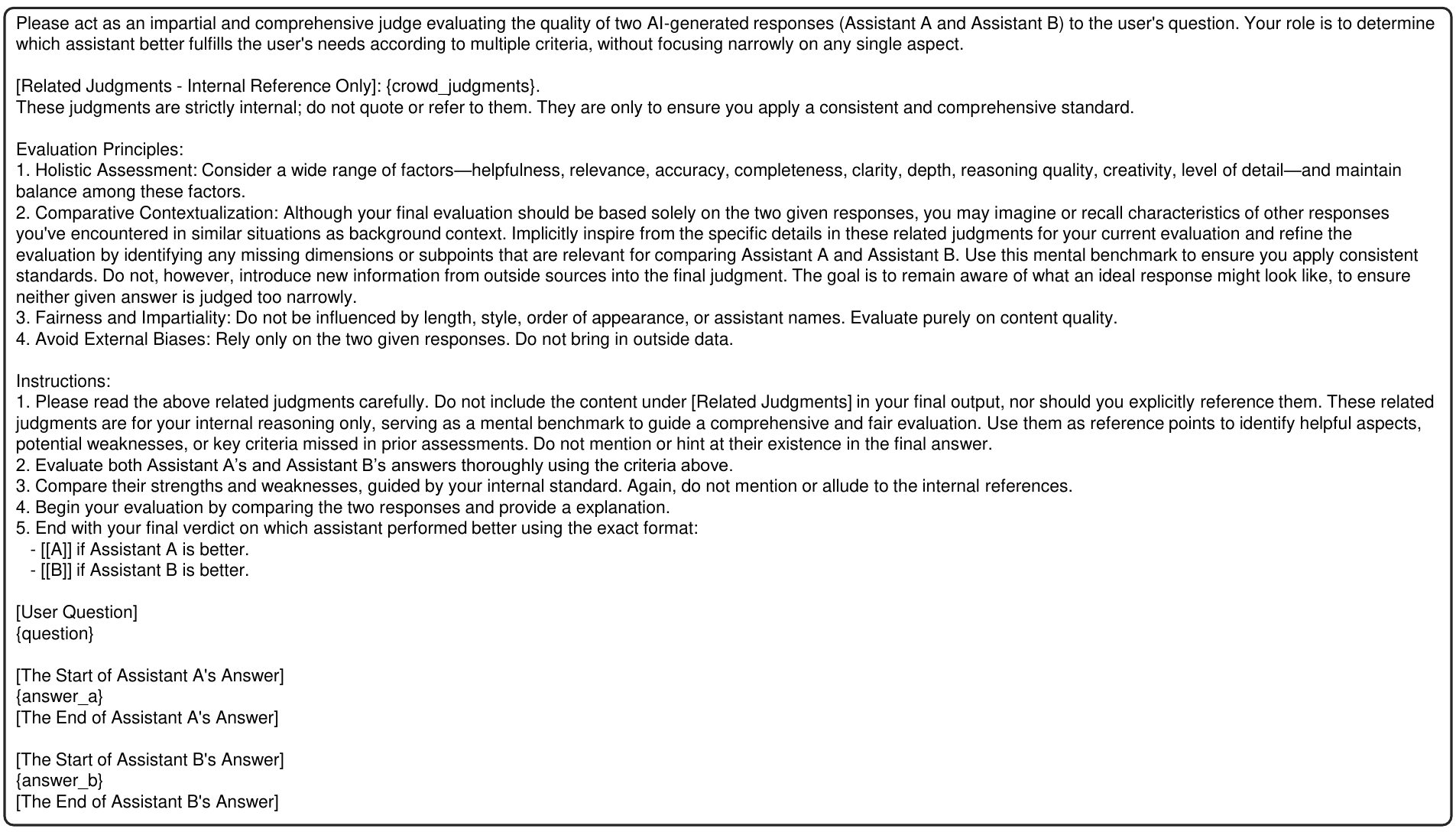}
  \caption {Prompt of Our Method.}
  \label{fig:prompt_our}
\end{figure*}

\begin{figure*}[!t]
  \includegraphics[width=\linewidth]{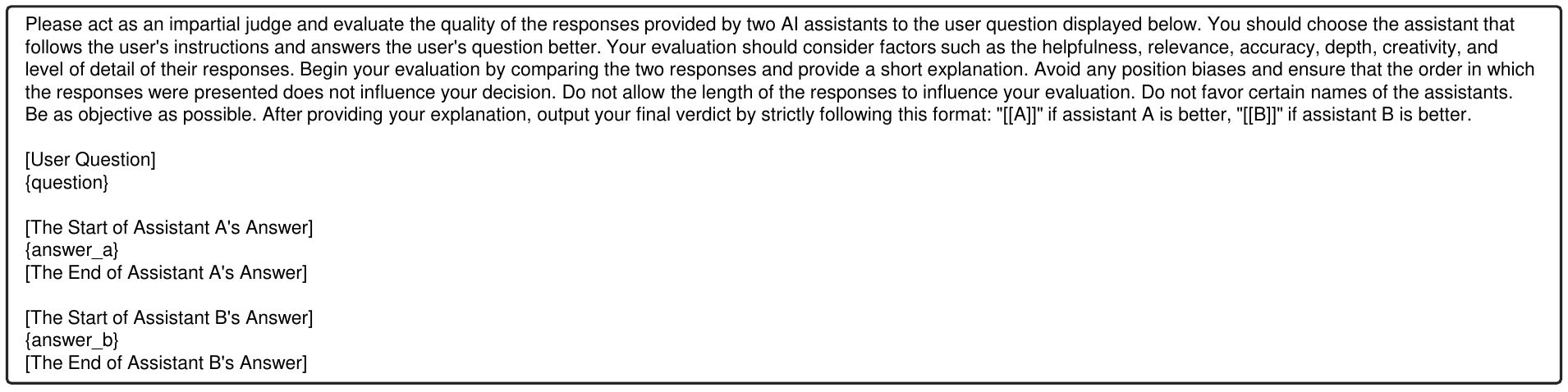}
  \caption {Prompt of Vanilla LLM-as-a-Judge.}
  \label{fig:prompt_vanilla}
\end{figure*}

\begin{figure*}[!t]
  \includegraphics[width=\linewidth]{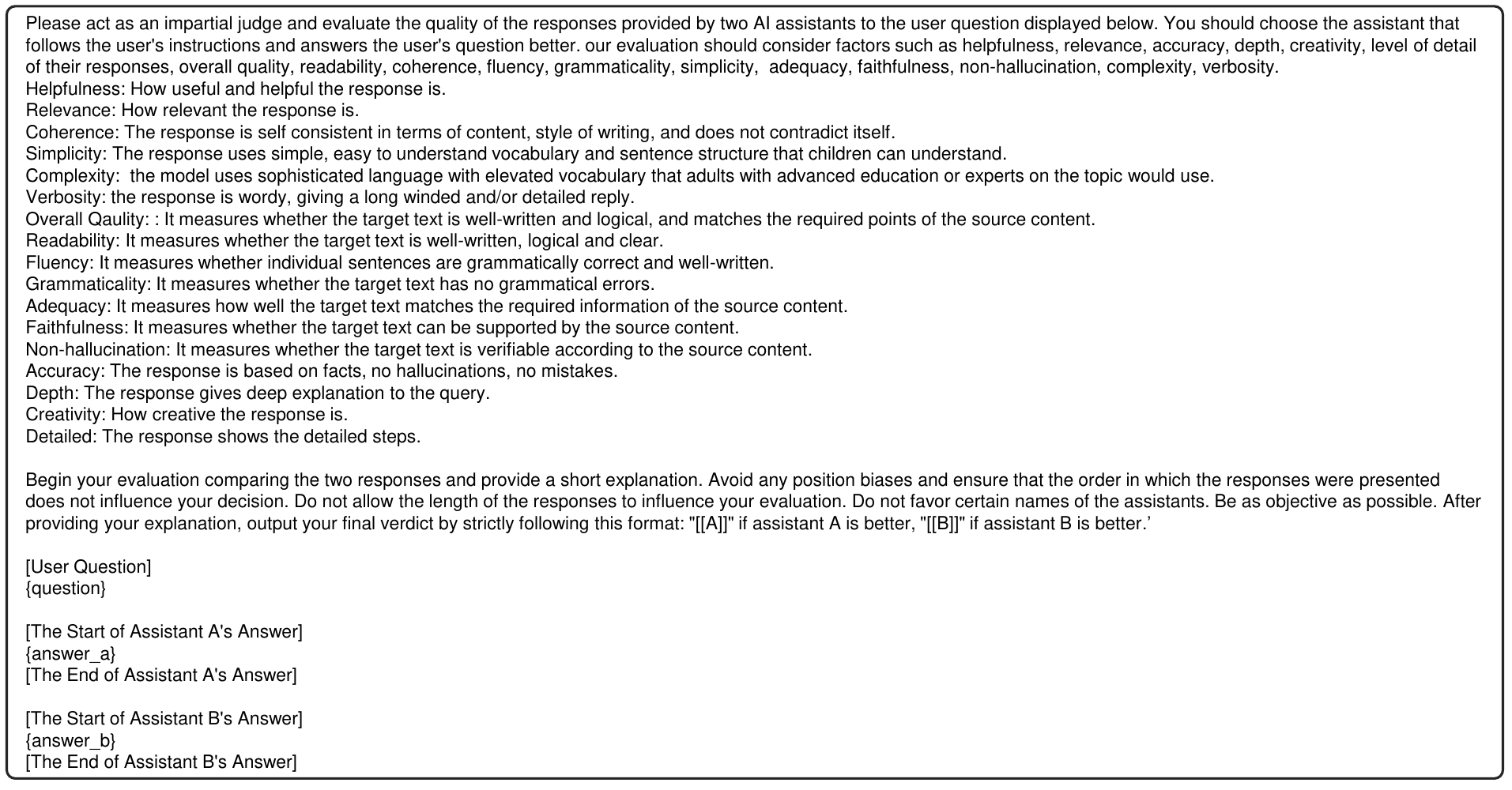}
  \caption {Prompt of 16-Criteria LLM-as-a-Judge.}
  \label{fig:prompt_16}
\end{figure*}

\begin{figure*}[!t]
  \includegraphics[width=\linewidth]{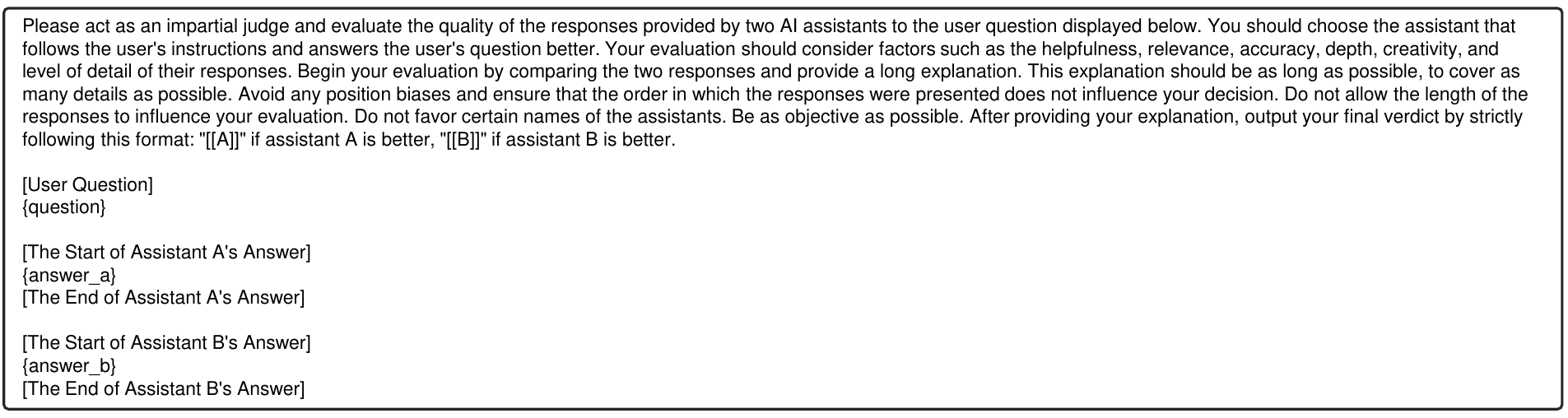}
  \caption {Prompt of LongPrompt LLM-as-a-Judge.}
  \label{fig:prompt_long}
\end{figure*}

We provide the prompt we used in this work for the experiment, as depicted in Figure~\ref{fig:prompt_our}. For Vanilla LLM-as-a-Judge (Figure~\ref{fig:prompt_vanilla}), we deployed the prompt designed in \textsc{MTBench}, which is widely deployed in many works, \eg, \textsc{RewardBench}. Notably, \textsc{HelpSteer2} specializes in 5 aspects, so we replace the \textsc{MTBench}'s aspects with these aspects when we test the method in \textsc{HelpSteer2}. Furthermore, we also present the prompts of baselines: \textit{LongPrompt}(Figure~\ref{fig:prompt_long}) forces the CoT as long as possible; \textit{16-Criteria} (Figure~\ref{fig:prompt_16}) incorporates 16 criteria and corresponding descriptions, which are designed in \citet{hu2024arellm} and \citet{wang2024helpsteer}.

\section{Testing Preference Benchmark}
\label{sec:testing}

\subsection{Preference Benchmarks}
\label{subsec:benchmark}
As shown in Table~\ref{tab:benchmarks}, we give a brief introduction to preference benchmarks. Each of these benchmarks has its own strengths; thoroughly testing all of them and averaging the results is a reliable way to evaluate the method. Notably, we randomly sampled $1$K cases from the training split of \textsc{EvalBias} since the size of the test split is 80 items, which is too small. 
\begin{table}[!tp]
  \centering
  \resizebox{0.48\textwidth}{!}{
  \begin{tabular}{lcc}
    \hline
    \textbf{Benchmarks} & \textbf{\textsc{Size}} & \textbf{\textsc{Focus}} \\
    \midrule
    \textit{RewardBench} &$2,985$& \makecell{It covers multiple  scenarios, including\\ Chat, Chat-Hard, Safety, and Reasoning.}\\
    \midrule
    \textit{HelpSteeer2} &$519$&\makecell{It provides multiple fine-grained dimensions\\ for evaluation, like Helpfulness, Coherence,\\ Correctness, Complexity, Verbosity.}\\
    \midrule
    \textit{MTBench Human} &$2,665$& \makecell{It provides multi-turn conversation for evaluation,\\ and we filter the samples whose outcome is ``Tie''.}\\
    \midrule
    \textit{JudgeBench} &$350$& \makecell{It focuses on challenging response pairs spanning\\ knowledge, reasoning, math, and coding}\\
    \midrule
    \textit{EvalBias} &$1,000$&\makecell{It tests the robustness of judges on\\ various scenarios containing evaluation biases.} \\
    \bottomrule
  \end{tabular}
  }
  \caption{The brief description of Preference Benchmarks for testing.}
  \label{tab:benchmarks}
\end{table}

\subsection{The Implementation of Generating Crowd Judgments}
\label{subsec:generate_crowd}

To generate crowd judgments, we produce a wide range of diverse responses. We employed several API-accessible and open-source LLMs to generate these responses based on the given instructions. Since diversity is crucial, we did not limit ourselves to only the most powerful models. Specifically, we used the following LLMs: \textit{Qwen-2.5-0.5B-Instruct}, \textit{Qwen-2.5-1.5B-Instruct}, \textit{Qwen-2.5-3B-Instruct}, \textit{Qwen-2.5-7B-Instruct}, \textit{Llama-3.2-1B-Instruct}, \textit{Llama-3.2-3B-Instruct}, \textit{Llama-3.1-8B-Instruct}, \textit{Mistral-Nemo10-Instruct-2407}, \textit{Mistral-7B-Instruct}, \textit{GPT-4o-mini}, and \textit{GPT-4o}. Additionally, we applied two temperature settings ($0.7$ and $1.0$) for each model. In principle, greater diversity in models and temperature configurations leads to improved performance.

Based on these crowd responses, we deployed the vanilla LLM-as-a-Judge to judge each crowd response with candidate response A/B separately using the judge LLM. 

\subsection{The Implementation of Baselines}
\label{subsec:implemenatation_baseline}

For maj@16 and agg@16, we modify the temperature setting to $1.0$ to promote more diversified responses. For other inferences in baselines, we set a unified temperature as $0.1$.

\subsection{The Implementation of Selection and Processing}
\label{subsec:implementation_selection}

For the selection strategy, we adopted ``Criticize Selection'' by choosing the crowd judgment where the outcome indicates that response A/B loses. For ``Outcome Removal Processing,'' we used \textit{GPT-4o-mini} to eliminate the outcome segment from the judgment with a temperature of $0$. The prompt is: 
\begin{quote}
``\textit{You are a helpful assistant. Specifically, I will provide you with the text quality judgment from an LLM-as-a-Judge evaluation of the responses from two AI assistants to an instruction. I need you to remove the final conclusion segments and only remain the evaluation analysis segments as soon as possible. ONLY OUTPUT the processed judgment.} ''

``\textit{*Judgment:* \{judgment\}}''
\end{quote}

\subsection{The Implementation of Inference}
\label{subsec:implementation_inference}

We tested our method on multiple LLMs-as-Judge, including \textit{GPT-4o} (2024-08-06), \textit{Qwen 2.5-7B-Instruct}, \textit{Qwen 2.5-32B-Instruct}, \textit{Qwen 2.5-72B-Instruct}, and \textit{Llama 3.3-70B-Instruct}. We found that reliability and consistency of evaluation can be balanced when temperature$=0.1$.

\section{Distilling CoT for Training Judge}
\label{sec:distilling4training}

\subsection{Distilling Preference Source}
\label{subsec:distilsource}
We chose the TULU3-Preference-Mixture~\footnote{\url{https://huggingface.co/datasets/allenai/llama-3.1-tulu-3-8b-preference-mixture}} as the preference data source. Specifically, we prompt the LLM-as-a-Judge to generate a CoT using the given instruction along with the chosen-rejected response pairs as input. Additionally, we experiment with two training sizes: random samples of $10$K and $30$K examples.

\paragraph{Distilling Inference.} We use the \textit{GPT-4o} as the Judge to produce the CoT, and the temperature setting is $0.1$. 

\subsection{The Implementation of Training Judge}
\label{subsec:implementation_trainingjudge}

\paragraph{Base Models.} To verify the generality of our method in Distilling CoT, we fine-tuned the preference data and corresponding CoT judgment in base LLMs: \textit{Qwen 2.5-7B-Base} and \textit{Llama 3.1-8B-Base}.

\paragraph{Training Setting.} We trained the Base LLM with a \textit{context length}$=4,096$, \textit{epochs}$=3$, \textit{batch size}$=128$,and \textit{learning rate}$=2e^{-5}$.

\section{SFT Data Selection}
\label{sec:sft_data_selection}

\subsection{Synthetic Response Pool for Selection}
\label{subsec:synthetic}
To enhance the challenge and realism of the SFT Data Selection, we chose four LLMs with similar general generation capabilities as the base models for synthesizing responses. These are: \textit{GPT-4o}, \textit{DeepSeek-v3}, \textit{Claude-3.5-Sonnet}, and \textit{Qwen 2.5-72B-Instruct}. For inference, we set the temperature parameter to 0.7. We generate four responses for each instruction to serve as the basis for subsequent selection. The base instruction queries we used are two pools: LIMA and TULU3-SFT. LIMA~\footnote{\url{https://huggingface.co/datasets/GAIR/lima}} contains 1,000 instructions, which are regarded as high-quality; TULU3-SFT~\footnote{\url{https://huggingface.co/datasets/allenai/tulu-3-sft-mixture}} contains $93.9$K instruction-response pairs, and we randomly sampled $10$K instructions as the query. The latter is the latest released multilingual dataset.

\subsection{The Implementation of Rejection Sampling}
\label{subsec:implementation_rejection}

Under the vanilla LLM-as-a-Judge approach, we perform pairwise comparisons among four responses, awarding a score of $+1$ to the winner of each matchup. After all comparisons, the response with the highest total score is selected. Building on this, our method incorporates the remaining two responses as ``crowd responses'' during each evaluation, allowing us to gather additional crowd judgments.

\paragraph{Base Judge Model.} The base judge model is \textit{GPT-4o}, and the temperature is set as $0.1$.

\subsection{The Implementation of Training SFT}
\label{subsec:implementation_trainingsft}

\paragraph{Base Models.} To verify the generality of our method in SFT data selection, we fine-tuned the instruction and selected response in base LLMs: \textit{Qwen 2.5-7B-Base} and \textit{Llama 3.1-8B-Base}.

\paragraph{Training Setting.} We followed the common setup for SFT, with a \textit{context length}$=2048$, \textit{epochs}$=3$, \textit{batch size}$=128$,and \textit{learning rate}$=2e^{-5}$.

\section{Inference Scaling}
\label{sec:infer_scal_appendix}

The ``Vanilla'' setup has no crowd judgments, ``1'' includes a single judgment, and even-numbered settings split judgments evenly between A and B. We use \textit{GPT-4o} as the judge and sample three times per setting to obtain the average result.

\section{CoT Comparison}
\label{sec:cot_comp_appendix}
\subsection{Key Points Extraction}
We use the Key points statistic to measure the richness of the CoT. Firstly, we use the \textit{GPT-4o-mini} to summarize the CoT to aspects and corresponding sub-points. The summarization prompt is 

\begin{quote}
    \textit{``Extract the key evaluation aspects and detailed points mentioned in the text below. List the aspects and points in a strictly structured format:''}
    
    \textit{``Example Input: `The response is accurate but lacks creativity. It includes factual details but misses key arguments.' ''}
    
    \textit{``Example Dictionary Output:''}
    \textit{``- Aspect: Accuracy ''}
    \textit{``  - Sub-point: Includes factual details ''}
    \textit{``  - Sub-point: Misses key arguments ''}
    \textit{``- Aspect: Creativity ''}
    \textit{``  - Sub-point: Lacks originality''}

    \textit{``**Input**:''}
\end{quote}

When we generate the summarized dictionary parsed output, we can get the total number of key points of each CoT.

\subsection{Coverage Rate Compuataion}
\label{subsec:coverage_appendix}
An attention-based approach computes mapping weights linking output tokens to input tokens. Interpretability research~\citep{bibal2022attention,vig2019multiscale} uses these weights to assess which input tokens influence the output. Our goal is to quantify how thoroughly CoT evaluates details in the target text, and attention-based computation provides a precise method for doing so.

Naturally, we used the \textit{bart-base}~\footnote{\url{https://huggingface.co/facebook/bart-base}} to compute the cross-attention between the target text and the generated CoT. We extract the cross-attention weights from the last layer of the decoder. By averaging these weights across attention heads and applying a threshold$=0.3$, it calculates a coverage rate—the fraction of the target text’s tokens whose attention is above the threshold from the CoT.

\end{document}